\documentclass[lettersize,journal,anonymous]{IEEEtran}
\usepackage{multirow}
\usepackage{subcaption}
\pdfoutput=1
\usepackage{amsmath,amsfonts}
\usepackage{algorithmic}
\usepackage{algorithm}
\usepackage{array}
\usepackage{textcomp}
\usepackage{stfloats}
\usepackage{url}
\usepackage{verbatim}
\usepackage{graphicx}
\usepackage{cite}
\begin{document}

\title{FedGAI: Federated Style Learning with Cloud-Edge Collaboration for Generative AI in Fashion Design}

\author{Mingzhu~Wu,~Jianan~Jiang,~Xinglin Li,~Hanhui~Deng,~Di~Wu,~\IEEEmembership{Member,~IEEE}
\IEEEcompsocitemizethanks{
\IEEEcompsocthanksitem All authors are with the National Engineering Research Center for Robot Visual Perception and Control Technology, Hunan University, Changsha, China. (E-mail: wumz@hnu.edu.cn; jiangjn22@hnu.edu.cn; lixinglin@hnu.edu.cn; denghanhui@hnu.edu.cn; dwu@hnu.edu.cn)}}



\maketitle

\begin{abstract}
Collaboration can amalgamate diverse ideas, styles, and visual elements, fostering creativity and innovation among different designers. In collaborative design, sketches play a pivotal role as a means of expressing design creativity. However, designers often tend to not openly share these meticulously crafted sketches. This phenomenon of data island in the design area hinders its digital transformation under the third wave of AI. In this paper, we introduce a Federated Generative Artificial Intelligence Clothing system, namely FedGAI, employing federated learning to aid in sketch design. FedGAI is committed to establishing an ecosystem wherein designers can exchange sketch styles among themselves. Through FedGAI, designers can generate sketches that incorporate various designers' styles from their peers, drawing inspiration from collaboration without the need for data disclosure or upload. Extensive performance evaluations indicate that our FedGAI system can produce multi-styled sketches of comparable quality to human-designed ones while significantly enhancing efficiency compared to hand-drawn sketches.
\end{abstract}


\section{Introduction}
In the realm of fashion design, artistic creativity plays an indispensable role in the successful execution of design tasks. The correlation between creativity and the volume of novel ideas generated by designers has been explored in previous studies~\cite{srinivasan2010investigating}. Collaborative design proves instrumental in nurturing creativity and fostering innovation among designers. The distinctive perspectives and experiences of individuals in collaboration serve as mutual sources of inspiration, yielding a plethora of innovative design concepts and solutions\cite{zamenopoulos2018co}. 

Within the context of collaborative design, sketches assume a pivotal role as the initial expressions of design creativity. They function as conduits for ideas and concepts during the creative process, acting not only as visual representations but also as guides within the design journey. Sketches become a medium for communication and expression, contributing significantly to the collaborative design process~\cite{jonson2005design}. Designers, each possessing their unique styles and visual languages, can convey emotions, stories, and concepts in their distinctive ways~\cite{seivewright2012basics}. Throughout the collaborative process, sketches facilitate the sharing and exchange of ideas among designers with divergent styles, providing the foundational inspiration and direction needed for the overall design process. In recent years, fashion trends have experienced continual evolution, necessitating a perpetual update of clothing styles to align with the dynamic fashion landscape. This evolving demand from consumers and the market underscores the imperative for constant innovation in clothing design. Consequently, designers are compelled to adapt to consumer preferences for diverse styles by intensifying their commitment to innovation and incorporating a spectrum of aesthetics.

With the help of generative AI technology, the time required to learn diverse styles\cite{han2019clothflow,bhunia2022doodleformer} can be significantly reduced. This is highly beneficial for fashion designers who wish to focus on gaining inspiration for clothing design. FashionQ was developed as a creativity support tool to promote creative fashion design through three interactive visual models: fashion attribute detection, style clustering, and popularity forecasting~\cite{jeon2021fashionq}. In the image synthesis domain, generative adversarial networks (GANs)~\cite{yan2022toward, wu2023styleme} constitute powerful tools for fashion image generation. Furthermore, in the style transfer domain, GANs~\cite{ liu2021self, kim2022gra} play a great role in style features extracting and inferring. In comparison with another popular generative AI model Stable Diffusion~\cite{rombach2022high}, GANs also present lightweight features with fewer data and parameter requirements to facilitate deployment on the user end ({\em e.g.} MacBook or iPad) for creativity practice. 

Considering that AI models are typically trained on existing samples and often require designers to upload their sketches, design data and other information to cloud server to train the AI models, not all designers are inclined to share their data, and safeguarding the originality of works holds significant implications for fostering creative incentives\cite{cao2023comprehensive, zhang2023complete}. Especially, along with the application boom of generative AI in popularity over the past year, AI art generation tools have raised serious issues on infringing the work of artists by scraping their work from the internet to train AI models without their consent. Therefore, we are committed to designing a new sharing mechanism and cloud-edge collaboration platform with the motivation of preserving original creations, fostering collaborative work by automatically generating sketches, and achieving style fusion. To achieve these goals, following challenges need to be addressed:

\textbf{Challenge 1:} The exposure to various styles aids designers in comprehending diverse design languages and techniques, thereby fueling their creativity. How to generate collaborative sketches with different designers' styles in large quantities to inspire designers is an important and challenging task. 

\textbf{Challenge 2:} The utilization of generative AI typically necessitates the submission of a designer's works as training data and designers may not be willing to share their painstakingly crafted works. Then how to use generative AI technology to assist the designer's clothing design process on the premise that the collection of clothing designer works is not leaked?

\textbf{Challenge 3:} Considering the limited computational resources and memory on edge client devices (i.e. designer-side) and aiming to enhance GANs model speed on edge clients, as well as the computation speed of the central server in executing federated learning algorithms, achieving efficient and cost-effective multi-designer style fusion and data privacy preservation becomes a significant and challenging endeavor.
\begin{figure}[tb]
  \centering
  \includegraphics[width=\linewidth]{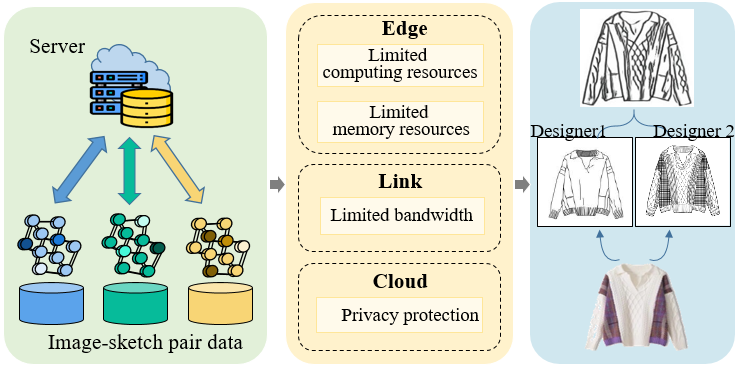}
  \caption{An overview of our FL-based style fusion system FedGAI.}
  \label{overview_FedGAI}
\end{figure}

\textbf{Our solutions}. To address these challenges, we introduce \textbf{FedGAI}, a \textbf{Fed}erated \textbf{G}enerative \textbf{AI} clothing system capable of automatically generating sketches with designer-specific style and generating sketches that incorporate different designers’ styles, while ensuring the privacy of the designer's intellectual property, as illustrated in Figure~\ref{overview_FedGAI}. Concerning \textbf{Challenge 1}, we employ federated learning with GANs to assist designers creatively while preserving data privacy. We aggregate parameters from locally deployed GAN models of multiple designers, distributing fused model parameters back to each designer. This enables designers to locally generate sketches with fused styles, aiding in the creative process. Regarding \textbf{Challenge 2}, Federated Learning (FL)\cite{konevcny2016federated,liao2023accelerating} meets the requirement well. As the FL framework only transfers model weights, preserving data privacy and reducing communication load~\cite{zheng2023autofed}. Despite GANs having sample generation quality not matching that of Diffusion and autoregressive models, they offer quick high-quality sample generation with lower resource demands, aligning with limited client computing resources, memory, and communication cost considerations. To address \textbf{Challenge 3}, we transmit only the GAN discriminator, significantly saving communication time as the generator's parameter size is typically larger than the discriminator's. We also employ GAN compression algorithms to reduce memory usage and improve runtime speed without compromising model accuracy on the designer's end. To the best of our knowledge, \textit{FedGAI is the first federated learning based generative AI framework to support intellectual property protection in creative design}. 

Our major contributions are summarized as follows:
\begin{itemize}
  \item We present a lightweight GAN for sketch generation to balance the generation quality and computing resource constraints on edge devices.

  \item We deploy GAN compression for local model acceleration, meanwhile alleviating the communication overhead of cloud-edge collaboration within federated learning. 
    
  \item We design a federated learning framework for privacy protection and style fusion in generative fashion design, with specific task allocation and coordination between cloud and edge.

  \item We implement FedGAI prototype with extensive evaluations. The results indicate FedGAI can learn personal style and perform on par with human designers in terms of style generation and fusion.
\end{itemize}

\section{Motivation}
\subsection{Intellectual Property Protection} In the era of digitization and the Internet, design works face heightened risks of intellectual property infringement due to easier dissemination and replication. Protecting design intellectual property is a paramount concern in creative industries\cite{cao2023breaking}. Research by Jiang et al.\cite{9679205} on legal cases related to design infringement has identified crucial design elements to mitigate infringement risks. Additionally, Casey and Amy\cite{fiesler2014copyright} explored how copyright awareness influences designers' willingness to share creative works online.
AI has emerged as a vital tool in the realm of design intellectual property protection. Digital watermarking technology\cite{8377886}, which embeds invisible information into digital media, enables designers to identify and monitor their works, enhancing tracking and verification efficiency. AI also facilitates the development of intelligent copyright management systems\cite{varaprasada2022design,kumar2019design} for secure digital media transfer between service providers and customers. While digital watermarking, blockchain, and related technologies deter direct copying and verify copyright information, they may have limited concealment capabilities~\cite{ChienMobiCom2023}. Once a work is public, it remains susceptible to uncontrollable forms of imitation. Numerous studies focus on tracing and protecting rights post-plagiarism, but they may not address the fundamental causes of potential infringement.

Federated Learning (FL)~\cite{konevcny2016federated} is a distributed machine-learning paradigm that transfers model weights instead of raw data to the central server. This approach ensures data privacy and reduces the communication load~\cite{zheng2023autofed}. To address ownership protection concerns in machine learning, Mohammed et al.~\cite{lansari2023federated} explored FL for knowledge property protection, focusing on FL watermarking. Yang et al.~\cite{yang2023federated} introduced FedIPR, a novel ownership verification scheme using watermarks embedded in FL models. Koushanfar and Farinaz~\cite{koushanfar2022intellectual} designed a covert communication framework to protect intellectual property by sharing information among local clients. Following these studies, our work aims to support fashion designers' intellectual property protection using generative AI technology in a privacy-preserving manner.

\subsection{Limited Resources of Edge Clients}
\begin{figure}[tb]
  \centering
  \includegraphics[width=\linewidth]{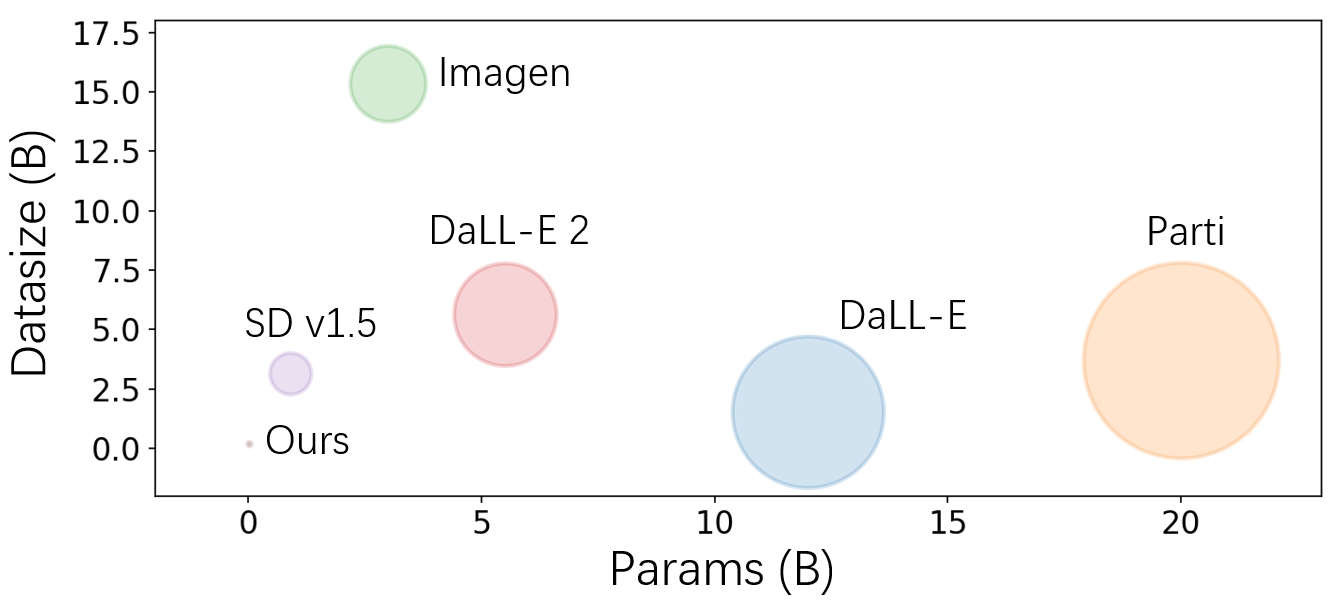}
  \caption{Comparison of different methods in model parameters and training data size.}
  \label{comparison_AR_DIFF_GAN}
\end{figure}
In recent years, Generative Artificial Intelligence (GAI) has gained significant prominence, attracting attention not only within the Information Technology domain but also across broader society. Notably, the advent of OpenAI's ChatGPT~\cite{lund2023chatting} has played a pivotal role in this. Generative AI enables the automated generation of substantial content within compressed timeframes, as opposed to content authored by humans~\cite{cao2023comprehensive}. In the realm of image generation, recent advancements have seen the emergence of autoregressive Transformer-based models like DALL-E~\cite{pmlr-v139-ramesh21a} and Parti~\cite{yu2022scaling}, as well as diffusion-based models like Imagen~\cite{saharia2022photorealistic} and Stable Diffusion~\cite{rombach2022high}. 
While the above models have achieved strong performance on image generation, they store all their knowledge inside the model, which tends to require a lot of parameters (e.g., 10B) and training data (e.g., 1B images)~\cite{yasunaga2022retrieval}. However, The model parameters for GANs (e.g., 10M) are generally at the level of tens of millions~\cite{tao2022df, xiao2022tackling}, which is clearly lightweight and convenient. The comparison of parameters and data required to autoregressive-based, diffusion-based, and GAN models can refer to Figure~\ref{comparison_AR_DIFF_GAN}. Furthermore, given the constrained computational resources and limited memory of terminal or edge devices, as well as the restricted bandwidth in communication links between cloud and edge, opting for a GAN with fewer parameters and lower memory requirements appears to be a more feasible choice.

\subsection{Limited Bandwidth of Cloud-Edge Link}
With the dramatic growth of generative AI models, data, and parameters \cite{fui2023generative, hacker2023regulating}, the application of Federated Learning (FL), as a collaborative cloud-edge inference mechanism, not only facilitates cooperation among multiple clients but also offers significant advantages in enhancing resource utilization efficiency, collecting new data for model training, reducing service latency, and mitigating user security and privacy threats \cite{huang2024federated, lomurno2022sgde}. 
Uploading and downloading model parameters may require significant bandwidth, especially in the era of Large Language Model \cite{kirchenbauer2023watermark}, when the number of devices increases or the volume of data is substantial. The communication bandwidth at the edge and cloud may become a bottleneck, affecting the speed of model updates. Therefore, we transmit only the GAN discriminator in the FL settings, significantly saving communication time as the generator's parameter size is typically larger than the discriminator's. Comparison of the aggregated parameter quantities at the cloud server in the FL environment using different generative models for varying numbers of clients can be referenced in Table \ref{tab:comparison_aggre_params}.  It can be observed that as the number of clients increases, the parameter count based on autoregressive models and diffusion models increases dramatically, while our method imposes the least burden on the cloud server.
\begin{table}[tb]
\caption{Comparison of parameters aggregated at cloud server based varying numbers of clients.}
\label{tab:comparison_aggre_params}
\scalebox{0.93}{
\renewcommand{\arraystretch}{1.25}
\begin{tabular}{llllll}
\hline
Model & Type & 5 clients & 10 clients & 15 clients & 20 clients \\ \hline
DaLL-E & Autoregressive & 60B & 120B & 180B & 240B \\
SD v1.5 & diffusion-based & 4.5B & 9B & 13.5B & 18B \\
StyleMe & GAN-based & 5.5B & 11B & 16.5B & 22B \\
Ours & GAN-based & 0.05B & 0.1B & 0.15B & 0.2B \\
Ours-D & GAN-based & 0.0075B & 0.015B & 0.0225B & 0.03B \\ \hline
\end{tabular}
}
\end{table}

\section{FedGAI System Design}\label{SSDI}
Our FedGAI system consists of three components: \textbf{(i)} the \textit{Lightweight GAN} for sketch generation (Section~\ref{3.1}) to efficiently and effectively learn fashion design styles from specific clothing images, \textbf{(ii)} the \textit{GAN Compression} for model acceleration (Section~\ref{3.2}) to mitigate computing and communication costs, and \textbf{(iii)} the \textit{FL} algorithm for style fusion and privacy protection (Section~\ref{3.3}) to integrate multiple design styles while preserving the users' creative intellectual property. 

\subsection{Lightweight GAN for Sketch Generation}\label{3.1}
\begin{figure*}[tb]
     \centering
        \includegraphics[width=0.75\linewidth]{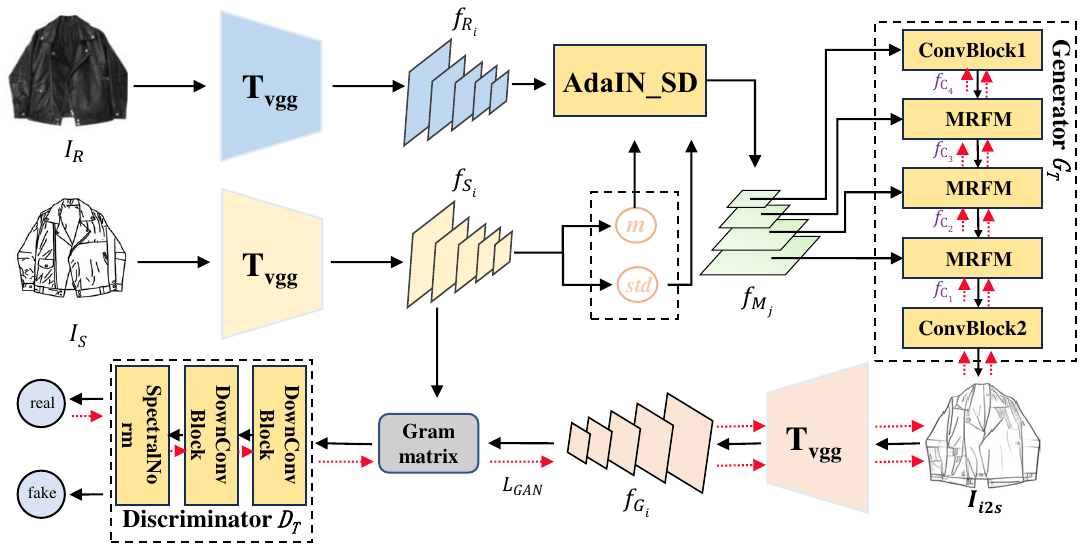}
        \caption{The lightweight GAN deployed on each client. Dashed lines indicate the flow of gradients for training the sketch generator.}
        \label{fig2}
\end{figure*}
\begin{figure}[tb]
    \centering
        \includegraphics[width=0.8\linewidth]{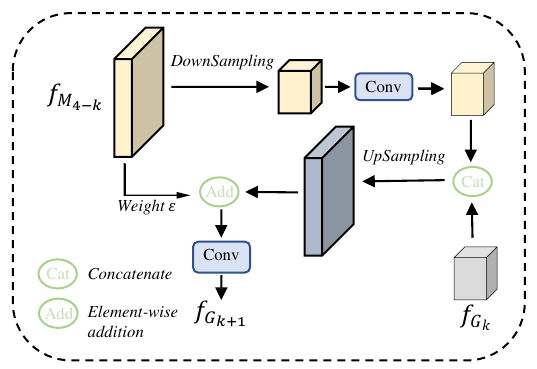}
        \caption{An overview of the \text{MRFM} module.}
        \label{fig3}
\end{figure}

As depicted in Figure~\ref{fig2}, we showcase the central component of our system: a lightweight GAN model tailored for sketch generation. This model is engineered to aid designers in swiftly producing hand-drawn sketches and fostering creative inspiration while reducing hardware requirements.

\subsubsection{Feature Encoder}
In the image generation process, our emphasis lies in reconstructing the image through simple contour features rather than complex semantic information. To achieve this, we opt for the VGG-16~\cite{simonyan2014very} model, known for its multi-level feature output capability, as the feature extractor. To extract the features of the inputs, we remove its final classification layer, focusing solely on multi-level feature extraction. To preserve local features and mitigate information loss caused by pooling layers, we choose the feature before the max-pooling operation as the output. Thus, each designer dataset's image-sketch pairs serve as input for sketch generation. Post-extraction via the VGG feature extractor, we obtained four feature outputs. Moreover, we fed the generated sketch back into the Encoder to ensure consistency between the generated sketch and the input image by our designed loss function. Hence, this process can be represented as: $f_{R_i}=T\left(I_R\right), f_{S_i}=T\left(I_S\right), f_{G_i}=T\left(I_{i2s}\right)$, where $i\in\{1,2,3,4\}$, $I_R$ denotes the image provided by the designer, $I_S$ represents the corresponding sketch paired with the designer's image, and $I_{i2s}$ signifies the corresponding sketch generated by the GAN deployed on the designer side.

\subsubsection{AdaIN\_SD}
To enhance the model's ability to capture designers' styles, we align sketch features with color image features. In the model training stage, we calculate the mean and variance of the four features extracted by \textit{Feature Encoder} for all sketch samples, denoted as $\mu\left(f_{S_j}\right)$ and $\sigma\left(f_{S_j}\right)$, $j \in \{1,2,3,4\}$. Here, $f_{S_j}$ represents the operation of computing the mean and standard deviation of the output features extracted by the \textit{Feature Encoder}. Notably, the normalization operations $\mu(\cdot)$ and $\sigma(\cdot)$ for $f_{S_j}$, i.e. are applied along the batch size, height, and width dimensions. After calculating the mean and standard deviation, $\mu(f_{S_j})$ and $\sigma(f_{S_j})$ are reshaped to match the size of $f_{R_j}$. As shown in Figure~\ref{fig2}, the extracted color image features $f_{R_j}$ and the processed sketch features $f_{S_j}$ with their mean and standard deviation are passed to the \textit{\text{AdaIN\_SD}} module for normalization. This process allows the model to adapt to distribution differences among various samples. Subsequently, the normalized features are input into the generator $G_T$.

\subsubsection{MRFM}
Our objective is to effectively harness existing features to enhance feature separation and generate high-quality sketches in a designer's style. To accomplish this, we introduce the Multi-Resolution Fusion Module (\text{MRFM}). As depicted in Figure~\ref{fig3}, the input features $f_{M_j}$ sequentially pass through the ConvBlock layer of the generator $G_T$. Initially, $ f_{M_4}$ enters the ConvBlock1 layer, and its output $f_{C_4}$ is amalgamated with $f_{M_3}$ before being forwarded to the subsequent layer, \text{MRFM}. Here, ConvBlock1 incorporates convolutional layers, 1-pixel padding, spectral normalization, batch normalization, and LeakyReLU. Specifically, \textit{MRFM} concentrates on the features extracted by the \textit{Feature Encoder} to enhance feature fusion by amalgamating features of the current level with those from the previous level. This fusion enables the generator to leverage a broader spectrum of information, encompassing features from various levels and dimensions, thereby improving the quality of generated outputs.

\subsubsection{Loss Function}\label{L_F}
The loss function of our proposed lightweight GAN model consists of three loss functions: \textit{Gram loss}, \textit{Adversarial loss}, and \textit{CLIP loss}.

\textit{Gram loss}. To ensure that the model captures detailed features such as texture, color, and spatial information of the image, we utilize Mean Square Error (MSE) loss at each level of the VGG encoder to constrain the Gram matrix ($\mathcal{G}$). 
\begin{equation}
\mathcal{L}_{\text {gram }}=\sum_{j=1}^4\left(\operatorname{MSE}\left(\mathcal{G}\left(f_{S_j}\right), \mathcal{G}\left(f_{G_j}\right)\right)\right)
\end{equation}

\textit{Adversarial loss}. We employ the adversarial loss to supervise the GAN generation process, accelerating the training of our models, which consists of the generator $G_T$ and the discriminator $D_T$. The $G_T$ and $D_T$ are alternately optimized using an adversarial objective to achieve their respective optimization goals.
\begin{equation}\label{eq:tea_LD}
\begin{gathered}
\mathcal{L}_{D_T}=-\mathbb{E}\left[\log \left(D_T\left(\mathcal{G}\left(f_S\right)\right)\right)\right]-\mathbb{E}\left[\log \left(1-D_T\left(\mathcal{G}\left(f_G\right)\right)\right)\right] \\
\mathcal{L}_{G_T}=-\mathbb{E}\left[\log \left(D_T\left(\mathcal{G}\left(f_G\right)\right)\right)\right]+\mathbb{E}\left[\left\|f_G-f_M\right\|^2\right] \\
\mathcal{L}_g=\mathcal{L}_{D_T}+\mathcal{L}_{G_T}
\end{gathered}
\end{equation}

\textit{CLIP loss}. We adopt the strong cross-modal expressive capability and global feature capability of the CLIP model~\cite{radford2021learning} and ViT
model \cite{dosovitskiy2010image}. To ensure the semantic consistency between the generated sketches and the reference hand-drawn sketches, we specifically utilize the last layer of features from the CLIP model, which are denoted as $f_g^{\text {final}}$ and $f_S^{\text {final}}$. Additionally, to preserve the coherence of global features like the contours and Lines of the generated sketches. We especially use the 4th layer features, which are denoted as $f_g^4$ and $f_R^4$. 
\begin{equation}
\mathcal{L}_{\text {CLIP }}=\operatorname{Cos}\left(f_g^{\text {final }}, f_S^{\text {final }}\right)+\operatorname{MSE}\left(f_g^4, f_R^4\right)
\end{equation}

In summary, the loss function is defined as:
\begin{equation}\label{eq:Shi7}
\mathcal{L}_{\text {GAN}} = \gamma_1 \mathcal{L}_{\text {gram}} + \gamma_2 \mathcal{L}_g + \gamma_3 \mathcal{L}_{\text {CLIP}}
\end{equation}
where $\gamma_1$ = 50.0, $\gamma_2$ = 1.0, and $\gamma_3$ = 25.0 are hyperparameters to achieve a balanced and stable training process.

\subsection{GAN Compression for Model Acceleration}\label{3.2}
\begin{figure}[tb]
    \centering
    \includegraphics[width=\linewidth]{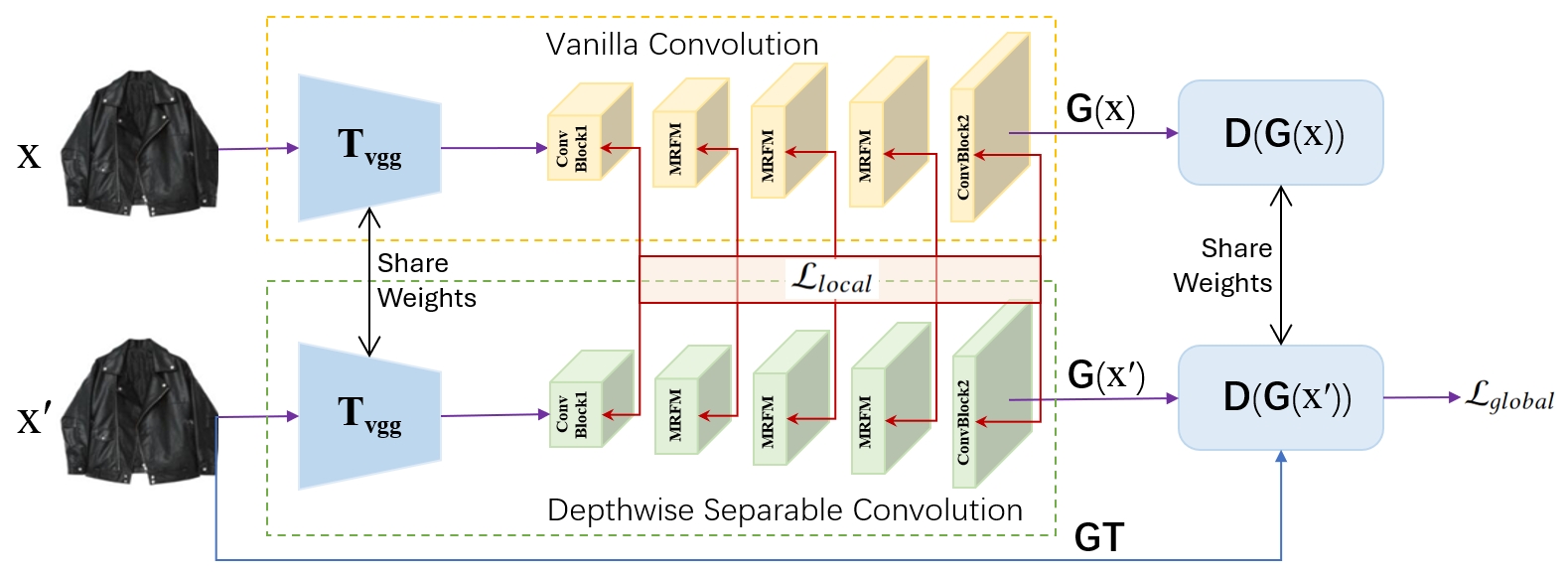}
    \caption{An overview of our GAN compression method.}
    \label{gan_compre}
\end{figure}
In the previous Section~\ref{3.1}, we present our core generation model, the \textit{Lightweight GAN}. In this section, we delve into further model compression techniques applied to the \textit{Lightweight GAN}. These techniques aim to alleviate the computational burden on the designer's equipment and expedite the sketch generation process. An overview of our GAN compression method is presented in Figure~\ref{gan_compre}.

\subsubsection{Preliminaries} 
GANs consist of two fundamental elements: the generator and the discriminator. Specifically, the generator is responsible for transforming the input into a synthetic image designed to deceive the discriminator. On the other hand, the discriminator's role is to differentiate between the outputs generated by the generator and real images. The generator and discriminator undergo iterative optimization processes through an adversarial term, each striving to attain its specific optimization goals. A widely used method for model compression is knowledge distillation, which was pioneered by Hinton et al.~\cite{hinton2015distilling} to allow a student classifier to mimic the output of its teacher. 

\subsubsection{Efficient Generator Design}
In knowledge distillation methods, the selection of an effective network structure significantly influences the generation performance of the model. The study conducted in \cite{li2020gan} demonstrated that merely reducing the number of channels in the teacher model does not lead to a compact student model.

In our method, the teacher model, as discussed in Section~\ref{3.1}, corresponds to the \textit{Lightweight GAN} introduced therein. The discriminator of the student model inherits its architecture from the discriminator of the teacher model. As can be seen in Figure~\ref{fig2}, the generator $G_T$ serves as the basis for the student model's generator, with the exception that the convolutional layers in the last two \textit{MRFM} modules of $G_T$ are replaced with depthwise separable convolutions~\cite{sifre2014rigid, howard2017mobilenets}, while the remaining structure remains unaltered. We find that using the depthwise separable convolutions also benefits the generator design in GANs.

\subsubsection{Local: Intermediate Feature Distillation.}
Conditional GANs like our proposed \textit{Lightweight GAN} typically yield a deterministic image as opposed to a probabilistic distribution. Consequently, extracting valuable insights, often referred to as dark knowledge, from the teacher's output pixels becomes challenging. This challenge is particularly pronounced in paired training scenarios where the output images generated by the teacher model essentially lack any supplementary information when compared to ground-truth target images. Relevant studies~\cite{li2020gan, liu2021content, li2021revisiting} demonstrate that simply replicating the output of the teacher model does not produce discernible improvement in such paired training settings.

To tackle this issue, we adopt an alternative approach by aligning the intermediate representations of the teacher generator. These intermediate layers, characterized by a larger channel number, offer a richer source of information. They empower the student model to gain additional insights beyond the final outputs. The distillation objective can be formally expressed as follows:
\begin{equation}
\mathcal{L}_{local}=\sum_{k=1}^K \operatorname{MSE}\left(f_{C_k}, f_{C_k}^{\prime}\right), K=4
\end{equation}
where $f_{C_k}, f_{C_k}^{\prime}$ denote the output of intermediate layer from teacher generator $G_T$ and student generator $G_S$, respectively.
\begin{figure*}[tb]
    \centering
    \includegraphics[width=0.88\linewidth]{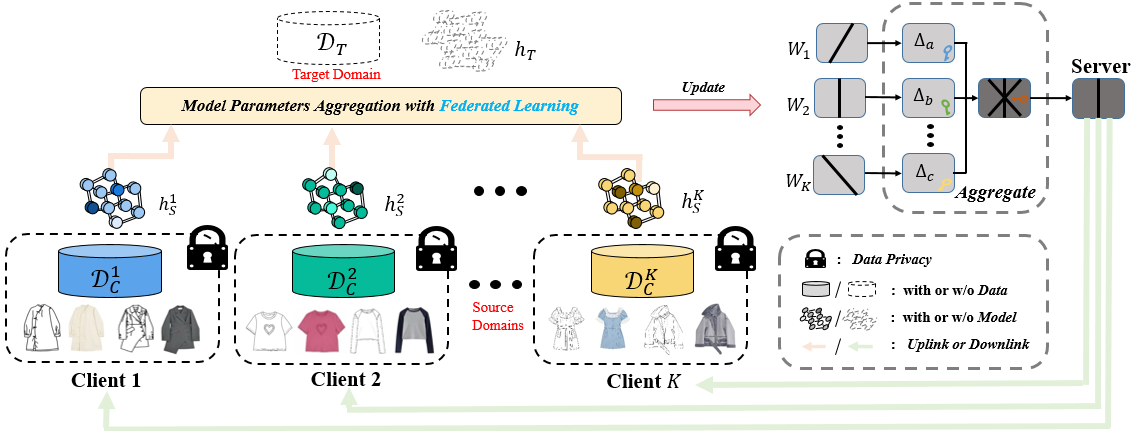}
    \caption{The generative federated AI system tailored for designer style fusion and intellectual property protection.}
    \label{fig1}
\end{figure*}

\subsubsection{Global: Paired Learning Loss.}
In the \textit{Lightweight GAN} that we proposed, the target domain corresponds to hand-drawn sketches provided by designers, while the source domain corresponds to colored images provided by designers. The GAN proposed in our work is trained using paired data. Our student GAN learning objectives can be summarized as follows:
\begin{equation}
\mathcal{L}_{global}=\mathbb{E}\left[\left\|f_G^{\prime}-f_S^{\prime}\right\|^2\right]
\end{equation}
where $f_G^{\prime}$ and $f_S^{\prime}$ represent the features of the student GAN-generated sketch and the features of the hand-drawn sketch, respectively, as illustrated in Figure 3, which can be referred to for $f_G$ and $f_S$. Although our goal is to compress the generator, the discriminator retains valuable knowledge about GAN learning, as it learns to identify the weaknesses of the current generator. Therefore, we employ the same discriminator architecture, initialize it with weights pre-trained by the teacher GAN, and fine-tune the discriminator alongside our compressed generator. The objective of the student GAN referred to Eq.~\ref{eq:tea_LD} is formalized as follows:
\begin{equation}
\mathcal{L}_{D_S}=-\mathbb{E}\left[\log \left(D_S\left(\mathcal{G}\left(f_S^{\prime}\right)\right)\right)\right]-\mathbb{E}\left[\log \left(1-D_S\left(\mathcal{G}\left(f_G^{\prime}\right)\right)\right)\right]
\end{equation}
where $D_S$ represents the student discriminator. Therefore, the full objective for our \textit{GAN Compression} is:
\begin{equation}\label{eq:GANC}
\mathcal{L}_{K D}=\mathcal{L}_{local}+\mathcal{L}_{global}+\mathcal{L}_{D_S}
\end{equation}

\subsection{FL for Privacy Protection and Style Fusion}\label{3.3}
The workflow, illustrated in Figure~\ref{fig1}, comprises three stages: designer-side style extraction, cloud server model parameter aggregation, and model parameter feedback and distribution.

Designer-side Style Extraction:
In this stage, multiple designers employ identical GAN models locally to extract style features from sketches drawn by different designers based on given color images. Only the discriminator network's model parameters are uploaded to the cloud server to minimize transmission time, network traffic, and computational load during cloud-based FL. $\mathcal{D}_C^K$ represents the dataset containing color images and sketches of the $K$-th designer, while $h_S^K$ represents the GAN model trained using the data of the $K$-th designer. Notably, the neural network model's structure, $h_S^K$, remains consistent across all designers, but GAN parameters are acquired through training on sketches drawn by different style-designing designers.

Cloud Server Model Parameter Aggregation:
In the cloud server model parameter aggregation stage, as seen in Figure~\ref{fig1}, there is no existence of data $D_T$ or model $h_T$. Utilizing our proposed federated algorithm, we can aggregate the discriminator model parameters $W_K$ from $K$ designers on the server side without the need to transmit designer data to the server.

Model Parameter Feedback and Distribution:
In the model parameter feedback and distribution stage, the model parameters aggregated via FL on the cloud server are simultaneously returned to each individual designer. By iteratively following the aforementioned stages, a set of model parameters encompassing various designer styles is aggregated and updated for each designer's model. Consequently, they can use their local dataset $\mathcal{D}_C^K$ to generate sketches that incorporate multiple designer styles.

\subsubsection{Problem Formulation and Assumptions}
The research is based on the assumption that the data distribution of each client is independently and identically distributed (i.i.d.). This means that the possibility of two individual clients having the same style is excluded. However, it is important to note that images and sketches from different clients are non-i.i.d., implying that they may exhibit diverse characteristics and style variations. 

Typically, the clients, i.e. designers, use relatively high-end laptops or desktop computers for sketching and designing purposes. The establishment of the FL framework is also predicated on the assumption that each individual client possesses sufficient computational capabilities to train both the shared and personalized models locally~\cite{silva2019federated}.

\subsubsection{Modified Update Strategy for Accelerating  Central Server-Side Computing}\label{batchnorm}
The system employs an efficient FL strategy. Similar to FedAvg~\cite{mcmahan2017communication}, it performs local updates and averages local model parameters. However, it assumes that the local models have BatchNorm layers and excludes their parameters from the averaging step. This operation is primarily aimed at saving computation time on the server side and enhancing the user experience for designers, who act as clients in the system.

As shown in Figure~\ref{fig1}, the algorithm's detailed procedure is as follows: The index for each designer is denoted by $k$, and the layer index of the GAN is denoted by $l$. The model parameters of the GAN are represented as $w_{0, k}^{(l)}$, the local update step size is denoted by $E$, and the total number of optimization rounds for both the server and edge clients is denoted by $T$. For each round $t$ ($t = 1, 2, ..., T$), the following operation is performed for each designer $k$ and each neural network layer $l$: $w_{t+1, k}^{(l)} \leftarrow S G D\left(w_{t, k}^{(l)}\right)$. In other words, stochastic gradient descent (SGD) is used to update the parameters of each layer of the generative adversarial network for each designer.

\subsubsection{Introducing the Regularization Term for Addressing Data Heterogeneity}\label{regul_term}
Considering the challenge of data heterogeneity in FL, where the local data of different participants (designers) comes from different distributions, it can significantly affect the overall model's final performance. The underlying reasons behind data heterogeneity and its impact on the model's expressive power are complex. One major issue arising from data heterogeneity is the dimension collapse of representations, severely restricting the model's ability to express and affecting the final performance of the global model. To address this, a FL regularization term called FedDecorr~\cite{shi2022towards} is introduced, and the loss function becomes:
\begin{equation}
\mathcal{L}_{Fed}=\mathcal{L}_{K D}+\beta \mathcal{L}_{\text {FedDecorr }}(w, X),
\end{equation}
where $\mathcal{L}_{K D}$ is the loss function of the compressed GAN referring to Eq.~\ref{eq:GANC}, $\beta$ is the coefficient of the FedDecorr regularization term, $w$ represents the model parameters of the GAN, and $X$ denotes the features of the generated sketches extracted by VGG, i.e. $f_{G}$.

The FedDecorr loss term, denoted as $\mathcal{L}_{\text {FedDecorr }}(w, X)$, is calculated as follows: $\mathcal{L}_{\text {FedDecorr }}(w, X)=\frac{1}{d^2}\|C M\|_{\mathrm{F}}^2$, where $CM$ is the correlation matrix. To compute $CM$, we can apply the z-score normalization~\cite{al2006data} on the representation vectors $f_{G_i}$ as follows: $f_{G_i}=\left(f_{G_i}-{f_{G_i}}_ \cdot \text { mean }(0)\right) / {f_{G_i}}_ \cdot \operatorname{std}(0)$
. Then, $CM$ can be computed as $C M=\frac{1}{\mathrm{~N}} * \operatorname{matmul}\left({f_{G_i}}_ \cdot t(), f_{G_i}\right)$, where $N$ is the batch size of $f_{G_i}$, $t()$ is the transpose operation and "matmul" denotes the matrix multiplication between tensors. 

\subsubsection{Client Selection for Controllable and Custom Style Fusion}\label{cli_select}
As illustrated in Figure~\ref{fig1}, $K$ designers participate in this process. Unlike conventional FL algorithms like FedAvg~\cite{mcmahan2017communication}, which blindly aggregate model weights from all clients, our approach is tailored to achieve a fusion of styles among designers. This aims to facilitate mutual reference and fulfill the inspirational needs of the designers. It's noteworthy that before applying FL algorithms, each designer locally deploys pre-trained GAN models capturing their distinct artistic styles. So, when a designer wishes to access another designer's artistic style, only the pre-trained GAN parameters specific to that designer (referred to as Designer A, for instance) are uploaded to the cloud server. These parameters are then transmitted to other interested designers aiming to understand Designer A's artistic style. Following this, other designers receive Designer A's style parameters and integrate them with their locally deployed GAN models, creating a parameter set that combines Designer A's style with their own. If the resulting style fusion is unsatisfactory, the process can be iterated by uploading Designer A's style parameters again, transmitting them to the remaining designers, and continuing until a convergence point is reached where designers are content with the style fusion effect.

This process remains consistent when designers wish to explore the styles of Designer B or any other designer. They can follow the same procedure outlined above for a seamless and collaborative style fusion experience. Furthermore, when designers aim to explore the fused styles of multiple designers, the process remains essentially the same. For instance, by uploading the parameters of Designers A, B, and E to the cloud server for aggregation using FL algorithms and subsequently distributing the fused style parameters to other edge designers, this iterative process continues until designers are satisfied with the fused styles of A, B, and E. The flexibility and controllability of this multi-designer style fusion customization depend on factors such as the computational power of the cloud server, FL algorithms, and communication bandwidth, among others.

\section{Performance Evaluation}\label{evaluation}
We initiated our evaluation by assessing the sketch generation capabilities of FedGAI in subsection~\ref{Experiment1}. We introduced three state-of-the-art sketch generation methods as baseline comparisons. Furthermore, we conducted ablation experiments on various FedGAI variants. Our evaluation spanned four key dimensions: sketch generation quality, model complexity, inference speed, and sketch generation speed.
We proceeded to evaluate FedGAI's sketch style fusion functionality in subsection~\ref{Experiment2}. To do this, we integrated three commonly used FL algorithms with the GAN model in FedGAI and employed them as baseline comparisons. We then compared FedGAI with these three baselines in terms of sketch fusion quality, sketch fusion stability, sketch generation speed, hyper-parameter experiment, and scalability experiment.

\subsection{Dataset}
Preceding the commencement of the experimental phase, a group comprising eight students specializing in fashion design, hailing from prestigious design institutions, was recruited to curate an extensive dataset comprising 
over 1000 pairs of clothing sketches. Each participant in this creative endeavor was tasked with creating approximately 200 distinct sketches, encompassing a wide spectrum of clothing types. 

These diverse garments encompass all common clothing categories, such as tops, skirts, vests, trousers, and more. The images also capture the stylistic characteristics of a diverse range of designers; for instance, some designers' sketches feature bold contour lines, while others emphasize delicate detailing. The collection includes pieces from various historical epochs, ranging from contemporary fashion ensembles to attire emblematic of ancient Asian traditions. To enrich the data for design reference, we also gathered sketches representing styles from different cultural backgrounds, such as Manchu, Mongolian, and other ethnic minorities.

To ensure the robustness and generalizability of the deep generative models and mitigate the risk of overfitting, a comprehensive and meticulously structured dataset augmentation strategy was employed. This augmentation strategy incorporates four distinct paradigms of image transformation, including cropping, horizontal flipping, rotation, and scaling. Furthermore, two distinct techniques were seamlessly integrated to enhance the fidelity of the sketch dataset, involving stochastic line erasure and line thickening.

\subsection{System Implementation}
All experiments are implemented based on PyTorch~\cite{10.5555/3454287.3455008} framework. In our FedGAI system, we deploy NVIDIA V100 GPU as the edge training and inference platform for personal sketch generation and style fusion on client side. The cloud server used for computation and fusion of different style hyperparameters is equipped with an Intel Xeon Platinum 8269CY CPU and 2GB RAM, with a network bandwidth of 1Mbps. Each participating client performs 11 local training epochs during each communication round, and a total of 11 rounds of FL communication are conducted. In each communication round, each client generates sketches that fuse the designated designer's style. The performance metrics we use are the average FID value across all clients aggregated on the cloud and the algorithm convergence time.

\subsection{Experiment Setup.}
\subsubsection{Baselines1.}
\label{Baseline_gan}
We executed comparative experiments in the realm of image-to-sketch generation, wherein we evaluated the efficacy of our proposed approach against three contemporary state-of-the-art methodologies. This assessment was carried out while adhering to identical preprocessing procedures.

\begin{itemize}
\item \textbf{USPS}~\cite{liu2020unsupervised} combines self-monitoring noise reduction objectives with attention modules to process inherent and specific abstract and stylistic changes in sketches.
\item \textbf{TOM}~\cite{liu2021self}
short for "Train Once and get Multiple transfers", the authors consider the sketch synthesis as an image domain transfer problem, where the RGB-image domain is mapped to the line-sketch domain.
\item \textbf{StyleMe}~\cite{wu2023styleme} utilizes an encoder, adaptive channel feature normalization, generator with channel attention module, loss function, and discriminator with CAM to efficiently generate personalized sketches from RGB images, achieving style-consistent and content-authentic sketch synthesis.
\end{itemize}

\subsubsection{Baselines2.}
To evaluate the performance of FedGAI, we compare FedGAI against three baselines:
\begin{itemize}
\item \textbf{FedAvg}~\cite{mcmahan2017communication}
is the first and perhaps the most widely adopted FL method. During training, all clients communicate updated local parameters to the central server and download the aggregated (i.e., averaged) global model for local training in the next round.
\item \textbf{FedYogi}~\cite{reddi2020adaptive} 
proposes a FL algorithm with an adaptive optimizer to address the drawbacks of FedAvg and other algorithms, which often struggle with parameter tuning and exhibit adverse convergence behavior.
\item \textbf{FedProx}~\cite{li2020federated} adds a proximal term to the loss function of local training to reduce the distance between the local model and the global model, hence addressing both system and statistical heterogeneity.
\end{itemize}

\subsubsection{Evaluation Metrics}
For the metrics to evaluate the quality of sketches generated by FedGAI's GAN module, we use: 
1) \textbf{FID} (Fréchet Inception Distance)~\cite{heusel2017gans}, which gauges the overall semantic realism of the synthesized images. The smaller the value, the higher the quality of the generated image. 2) \textbf{LPIPS} (Learned Perceptual Similarity)~\cite{zhang2018unreasonable}, which quantifies the perceptual dissimilarity between two images. The smaller the value, the higher the quality of the generated image. And for the metrics to evaluate FedGAI's GAN model complexity, we utilize: 1) \textbf{FLOPs} (Floating Point Operations)~\cite{molchanov2016pruning} is used to measure the complexity and inference speed of a neural network. 2) \textbf{Params} (Parameters) to calculate the number of parameters in a neural network.

\subsection{Generation Performance of FedGAI}\label{Experiment1}
We compare our model to the state-of-the-art methods mentioned in the subsection~\ref{Baseline_gan}. USPS~\cite{liu2020unsupervised}, TOM~\cite{liu2021self} and StyleMe~\cite{wu2023styleme}. Furthermore, We examined the performance of three model components in FedGAI, which are the \text{AdaIN\_SD}, the \text{MRFM}, and the CLIP loss in our loss function in subsection~\ref{L_F}, through two types of comparison tasks including sketch generation and style fusion.

\subsubsection{Qualitative experiments}
\begin{figure}[tb]
	\centering
	\includegraphics[width=1\columnwidth]{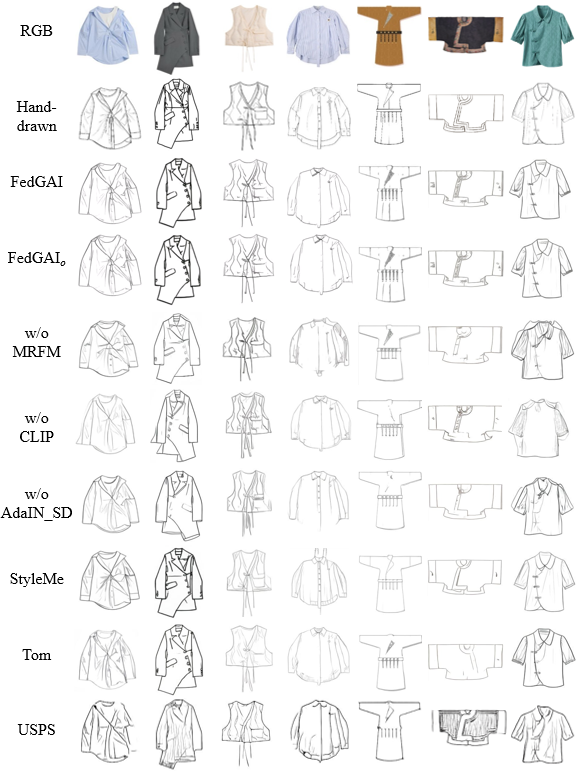}
	\caption{Sketch generation results of FedGAI, its variants, and other models.}
	\label{effect GAN}
\end{figure}

\begin{table*}[tb]
\caption{Sketch generation: Quantitative comparison to existing methods, bold indicates the best score}
\label{tab:skethgen}
\scalebox{0.84}{
\renewcommand{\arraystretch}{1.2}
\begin{tabular}{cccccccccccccccccc}
\hline
\multirow{2}{*}{Methods} & \multirow{2}{*}{FLOPs(G)} & \multirow{2}{*}{\begin{tabular}[c]{@{}c@{}}Params\\ (GB)\end{tabular}} & \multirow{2}{*}{\begin{tabular}[c]{@{}c@{}}Params-D\\ (MB)\end{tabular}} & \multicolumn{2}{c}{D1} &  & \multicolumn{2}{c}{D2} &  & \multicolumn{2}{c}{D3} &  & \multicolumn{2}{c}{D4} &  & \multicolumn{2}{c}{D5} \\ \cline{5-6} \cline{8-9} \cline{11-12} \cline{14-15} \cline{17-18} 
 &  &  &  & FID$\downarrow$ & LPIPS$\downarrow$ &  & FID$\downarrow$ & LPIPS$\downarrow$ &  & FID$\downarrow$ & LPIPS$\downarrow$ &  & FID$\downarrow$ & LPIPS$\downarrow$ &  & FID$\downarrow$ & LPIPS$\downarrow$ \\ \hline
USPS & 375.43 & 0.030 & 0.011 & 158.055 & 0.328 &  & 128.478 & 0.387 &  & 315.026 & 0.439 &  & 204.775 & 0.372 &  & 159.347 & 0.367 \\
Tom & 121.36 & 0.022 & 3.4 & 19.084 & 0.333 &  & 17.090 & 0.327 &  & 268.001 & 0.420 &  & 154.048 & 0.374 &  & 149.738 & 0.242 \\
StyleMe & 696.70 & 1.1 & 3.4 & 15.072 & 0.318 &  & 13.828 & 0.311 &  & 272.356 & 0.443 &  & 147.297 & 0.351 &  & 153.697 & 0.245 \\
w/o \text{MRFM} & 63.34 & 0.0161 & 1.5 & 16.143 & 0.356 &  & 15.755 & 0.301 &  & 288.059 & 0.458 &  & 155.910 & 0.405 &  & 150.660 & 0.297 \\
w/o CLIP & 18.19 & 0.0154 & 1.5 & 16.720 & 0.327 &  & 14.246 & 0.300 &  & 290.761 & 0.481 &  & 157.773 & 0.401 &  & 158.941 & 0.380 \\
w/o \text{AdaIN\_SD} & 18.19 & 0.0154 & 1.5 & 16.238 & 0.327 &  & 14.138 & 0.295 &  & 280.149 & 0.450 &  & 164.476 & 0.376 &  & 152.629 & 0.300 \\
$\operatorname{FedGAI}_{o}$ & 18.19 & 0.0154 & 1.5 & 13.345 & 0.299 & \textbf{} & 13.713 & 0.282 & \textbf{} & 281.021 & 0.444 &  & 155.356 & 0.374 &  & 146.116 & 0.290 \\
FedGAI & \textbf{5.44} & \textbf{0.0103} & \textbf{1.5} & \textbf{9.966} & \textbf{0.299} &  & \textbf{9.139} & \textbf{0.276} &  & \textbf{248.089} & \textbf{0.395} &  & \textbf{143.741} & \textbf{0.349} &  & \textbf{125.729} & \textbf{0.210} \\ \hline
\end{tabular}
}
\end{table*}

Figure~\ref{effect GAN} shows the effect of the FedGAI system and its variants to generate sketches. The FedGAI row represents the generated sketches of our FedGAI system, $\operatorname{FedGAI}_{o}$ row represents our FedGAI without GAN compression, w/o \text{MRFM} row represents our $\operatorname{FedGAI}$ without \text{MRFM} module, w/o \text{AdaIN\_SD} row represents our $\operatorname{FedGAI}$ without \text{AdaIN\_SD} module, w/o CLIP row represents our $\operatorname{FedGAI}$ without CLIP loss module. And for the column results in Figure~\ref{effect GAN}, the first column to the seventh column represents the first designer to the seventh designer, while the first and second rows depict the colored images created by each designer alongside their corresponding hand-drawn sketches.

The third row to the seventh row signifies the sketches generated by FedGAI and its variations, while the last three rows illustrate the generation outcomes of StyleMe, Tom, and USPS respectively. From Figure~\ref{effect GAN}, we can observe that the sketches generated by FedGAI exhibit a style that is remarkably close to hand-drawn sketches and has intricate details. In contrast, when examining StyleMe's generation of sketches for the fourth designer, it is evident that redundant lines are generated around the collar area. As for Tom, it can be noted that in generating sketches for the third, fourth, and sixth designers, there are noticeable deficiencies in detail and missing lines. However, USPS lacks attention to detail and missing lines.

In terms of the ablation experiments involving FedGAI, we can observe distinct deficiencies in the generated sketches for the fifth and seventh designers by  w/o \text{MRFM}, w/o \text{AdaIN\_SD}, and w/o CLIP. Specifically, w/o \text{MRFM} and w/o \text{AdaIN\_SD} exhibit missing lines in the sketches for the fourth designer, with w/o \text{MRFM} additionally generating overlapping lines. Furthermore, when generating sketches for the seventh designer, all three models—w/o \text{MRFM}, \text{AdaIN\_SD}, and w/o CLIP—introduce some disordered lines that deviate from the template. Concerning the fourth $\operatorname{FedGAI}_{o}$ row, we observe that $\operatorname{FedGAI}_{o}$ exhibits slightly better details in sketch generation for the third and fourth designers compared with FedGAI. This observation is further validated by Table.~\ref{tab:skethgen}, where $\operatorname{FedGAI}_{o}$ demonstrates lower FID and LPIPS values for the third and fourth designers compared to FedGAI.

\subsubsection{Quantitative experiments}

In Table.~\ref{tab:skethgen}, D1 to D5 denote the 1st to 5th designers, respectively. FLOPs quantify the complexity of the generator within the GAN model, Params refers to the overall parameter size of the entire GAN model, and Params-D denotes the parameter size of the discriminator within the GAN model. From the perspective of model complexity, it is indeed surprising that StyleMe exhibits significantly higher FLOPs and Params compared to other models. This high computational demand imposes a substantial barrier for designers, making it less user-friendly.
In contrast, as shown in Table.~\ref{tab:skethgen}, the FedGAI system, which incorporates GAN compression techniques, requires only 5.44 GFLOPs and has a model size of approximately 0.01 GB, offering a much more designer-friendly solution. Furthermore, when assessing the quality of sketch generation based on image-sketch pair data from five designers, it becomes evident that the top performers are FedGAI and $\operatorname{FedGAI}_{o}$, i.e. FedGAI without GAN compression. Specifically, in terms of FID and LPIPS, both models achieved the best scores based on data from the five designers. 
\begin{figure}[tb]
  \centering
  \begin{minipage}{0.45\textwidth}
    \centering
    \includegraphics[width=\linewidth]{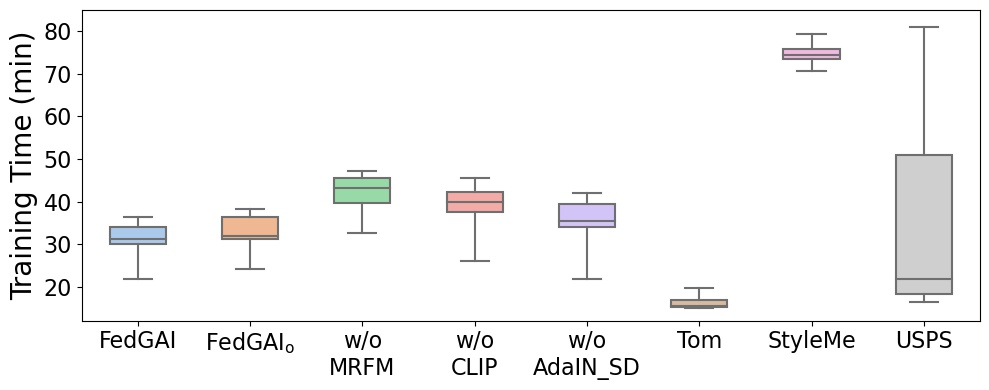}
    \caption{Training time of FedGAI, its variants, and other models based on designers' data.}
    \label{fig:training_time}
  \end{minipage}
\end{figure}

Training times for FedGAI and its variants, as well as other models on the data provided by our designers, can be observed in Fig~\ref{fig:training_time}. Notably, the most time-consuming model is StyleMe, with training times exceeding 70 minutes on the devices of our designers. This can be attributed to the substantial parameter size of StyleMe, as indicated in Table 1, which naturally leads to increased model complexity and, consequently, longer training times. And USPS has the most unstable training time.
Conversely, the model with the shortest training times is Tom, with training duration ranging from 10 to 20 minutes on designers' devices. This can be attributed to Tom's significantly smaller parameter size, as indicated in Table 1, resulting in lower model complexity. Additionally, Tom's FLOPs are less than 1/6th of StyleMe's, leading to faster inference speeds.
FedGAI, on the other hand, ranks as the second-fastest in terms of training time, with training times of around 30 minutes for these designers. None of the other variants match FedGAI's training speed. Although FedGAI is not the fastest, it excels in sketch generation quality, making it the top performer in this regard. In contrast, Tom tends to rank lower in sketch generation quality.

\subsection{Fusion Performance of FedGAI}\label{Experiment2}
\subsubsection{Style Fusion of Two Clients}
\begin{figure}[tb]
	\centering
	\includegraphics[width=\columnwidth]{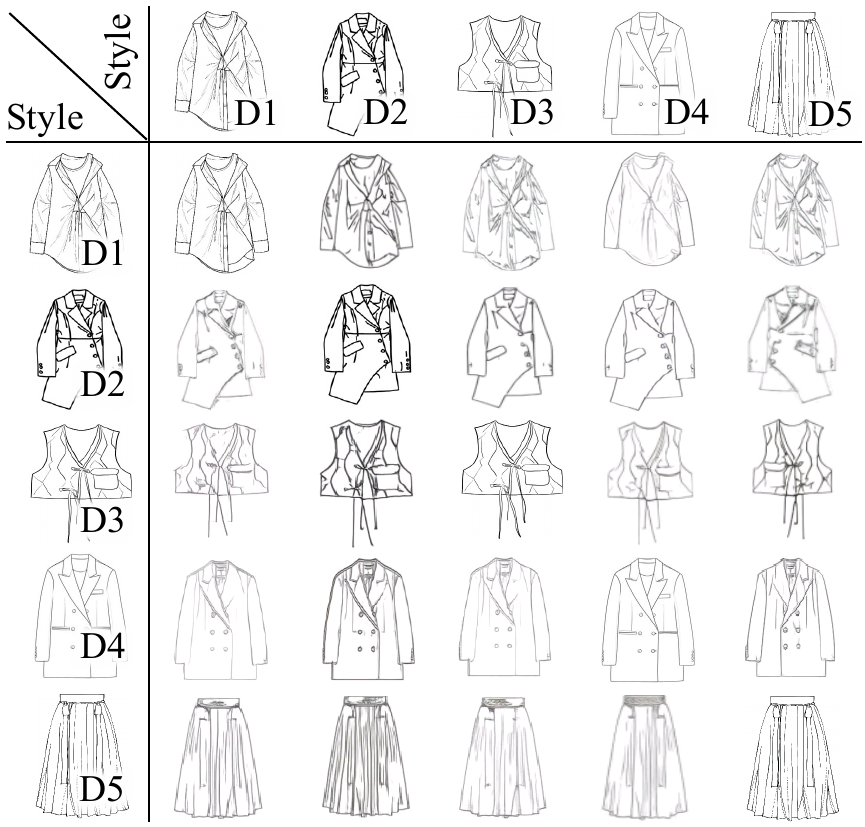}
	\caption{Sketch generation of style fusion using FedGAI federated AI system for five designers.}
	\label{FL_SM}
\end{figure}

The style fusion results of arbitrary two clients using FedGAI is depicted in Fig~\ref{FL_SM}. The first row and first column show hand-drawn sketch examples of five designers (D1 to D5), respectively. The second column displays sketches fused with the style of D1, such as the sketch in the third row, second column, which is fused with the style of D1 into D2's sketch. The third column exhibits sketches fused with the style of D2, for instance, the sketch in the second row, the third column, which incorporates D2's style into D1's sketch, and so forth. Consequently, the second through sixth columns present sketches fused with the styles of designers D1 through D5, respectively.

\subsubsection{FL Comparison on Style Fusion of Two Clients} 
\begin{figure}[htb]
  \centering
  \begin{minipage}{0.45\textwidth}
    \centering
    \includegraphics[width=\linewidth]{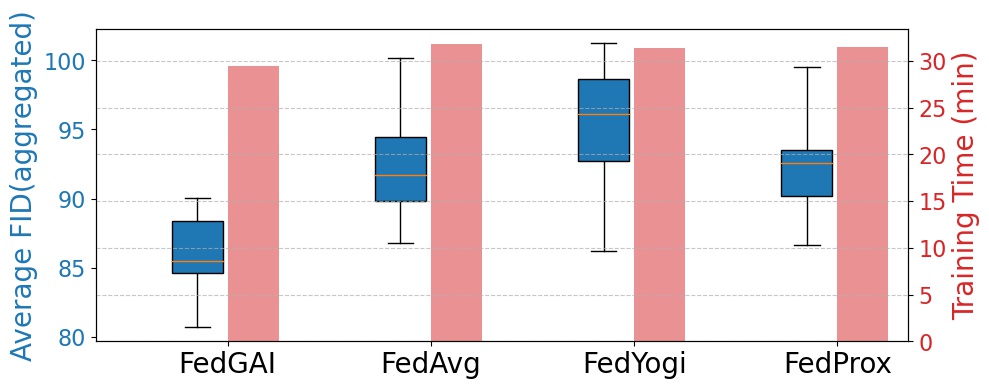}
    \caption{Comparing FedGAI with several baselines on FID and convergence time overhead.}
    \label{fig:FID_FT}
  \end{minipage}
\end{figure}

As for two designers' style fusion case with blue boxplots in Fig~\ref{fig:FID_FT}, within 11 rounds of FL communication, our FedGAI exhibits lower FID values compared to FedAvg, FedYogi and FedProx, which could be attributed to our personalized client selection strategy detailed in subsection~\ref{cli_select}. Additionally, it is noteworthy that our FedGAI shows minimal variation between its maximum and minimum FID values, displaying a relatively tighter distribution. This stability, compared with FedAvg, FedYogi and FedProx, possibly due to the regularization term introduced in subsection~\ref{regul_term}, which mitigates the oscillation of training patterns arising from data feature heterogeneity in GAN training. Furthermore, the red bar chart also demonstrates that FedGAI achieves the shortest convergence time among the four methods, which can be attributed to our FL training strategy outlined in subsection~\ref{batchnorm}, optimized within the aggregation parameter phase.

\begin{figure*}[htb]
    \begin{center}
        \begin{tabular}{cccc}
            \includegraphics[width=0.235\linewidth]{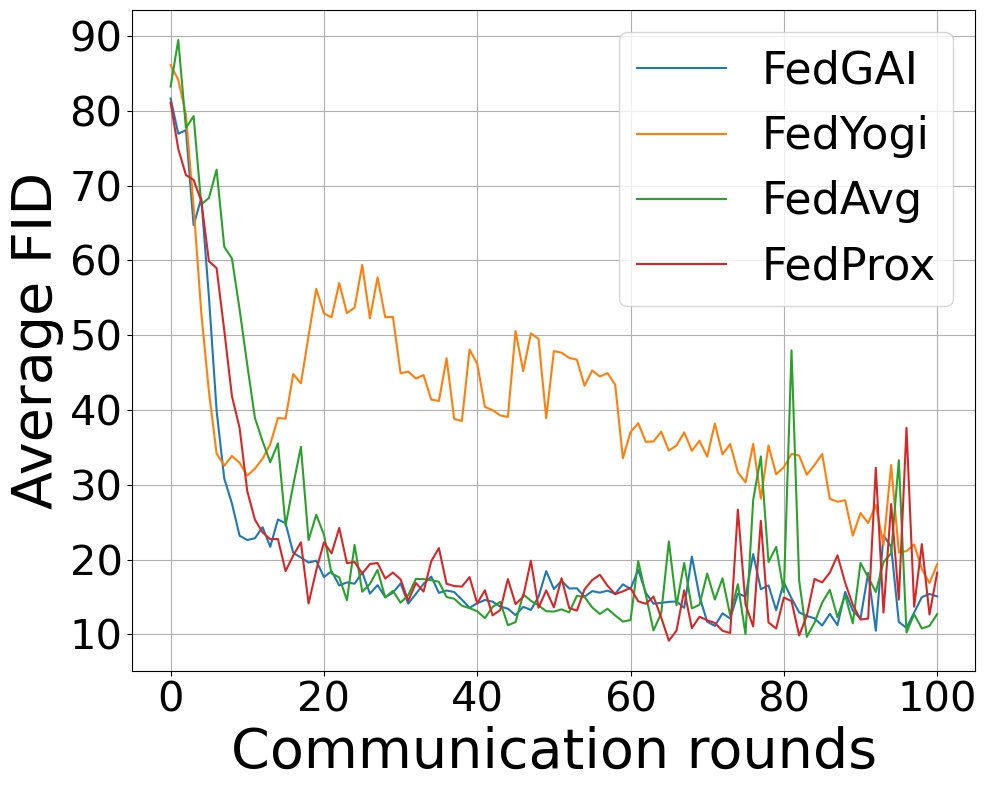}&
            \includegraphics[width=0.235\linewidth]{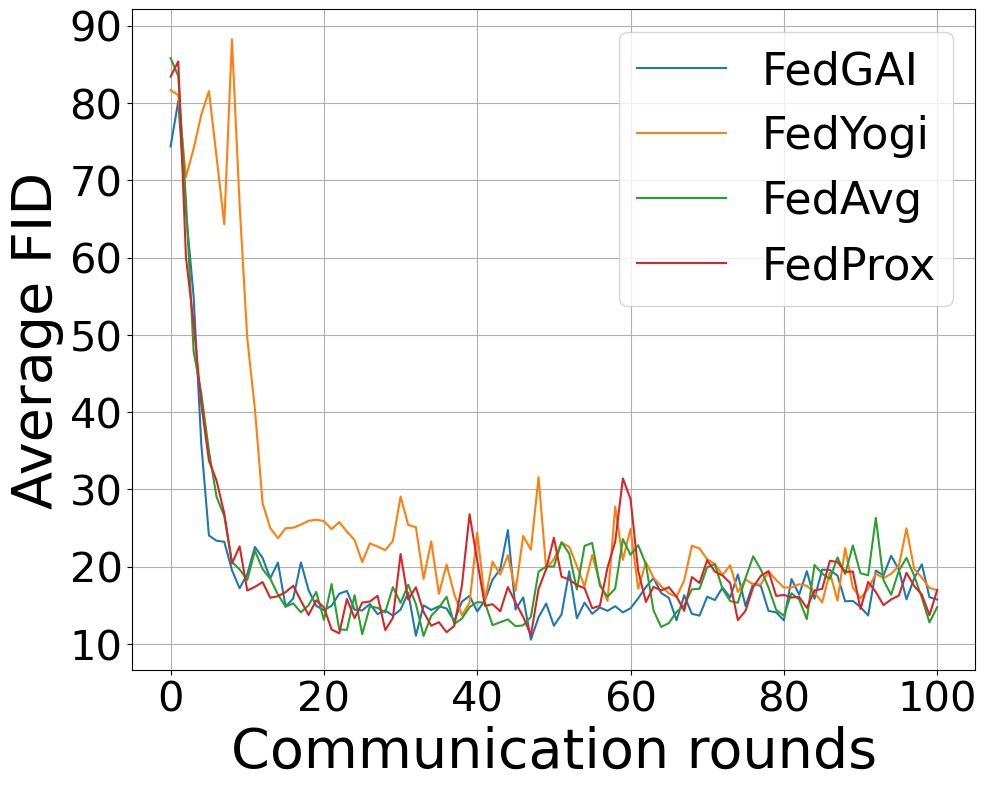}&
            \includegraphics[width=0.235\linewidth]{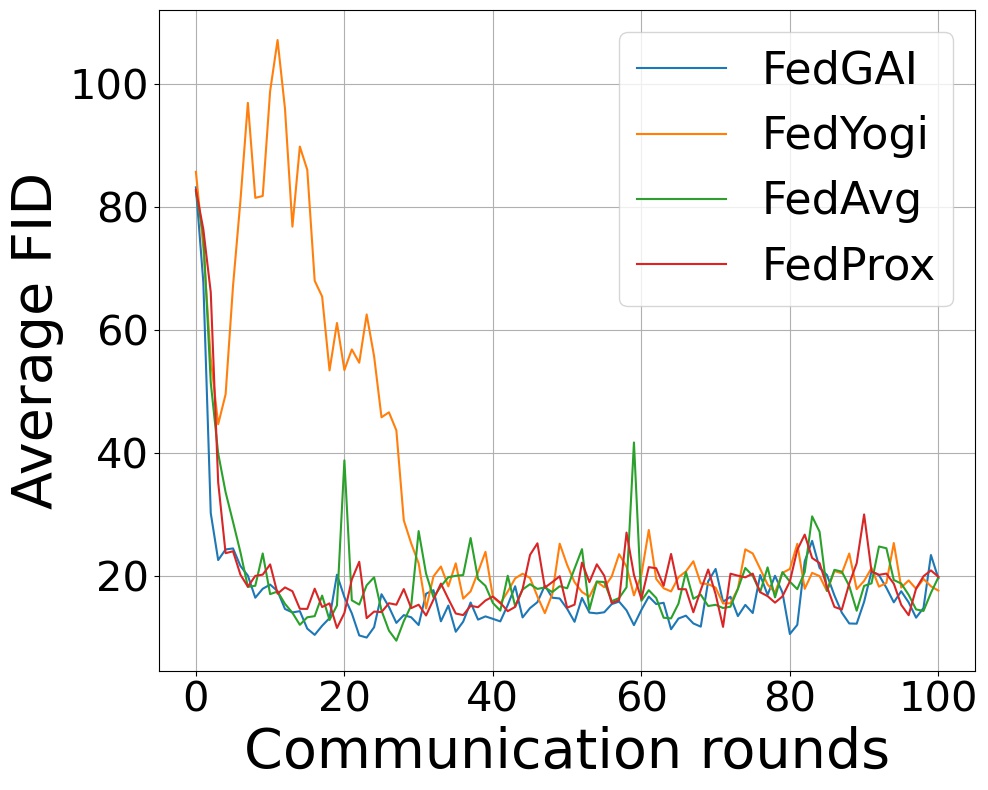}&
            \includegraphics[width=0.235\linewidth]{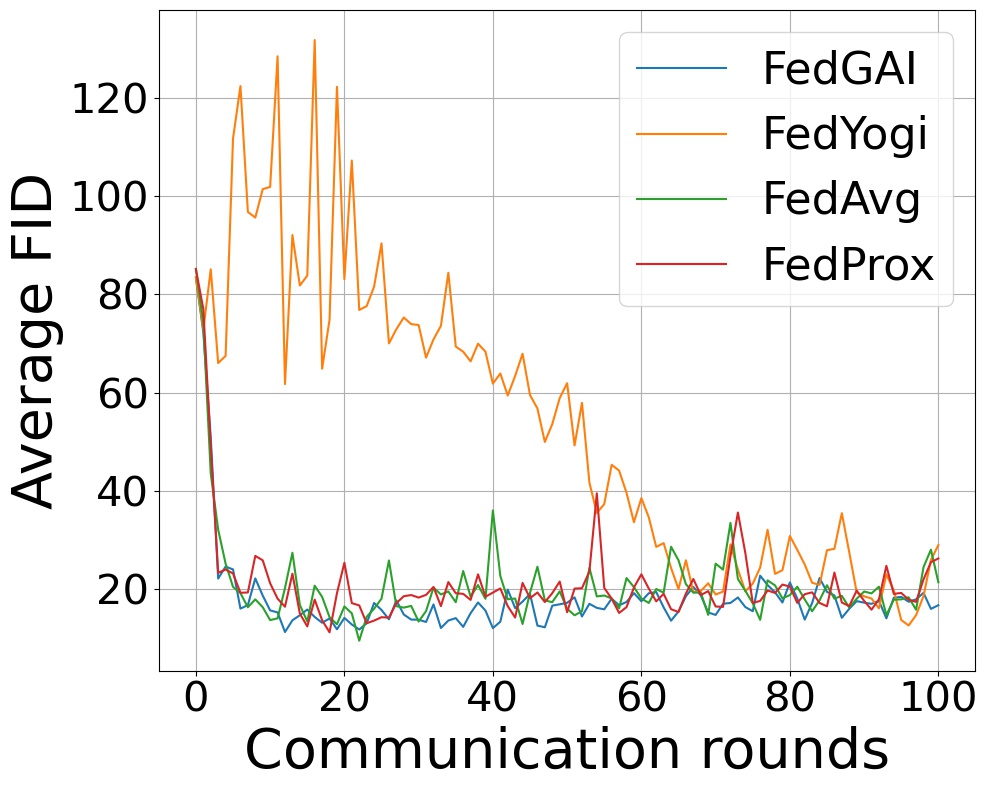}
            \\
            {\normalsize  (a) n\_iter=2}&
            {\normalsize  (b) n\_iter=5}&
            {\normalsize  (c) n\_iter=8}&
            {\normalsize  (d) n\_iter=11}
        \end{tabular}
    \end{center}
    \caption{The impact of local GAN model update iterations n\_iter on FL.}
    \label{fig:commun_n_iter}
\end{figure*}

\subsubsection{Hyper-parameter Experiment}
The hyperparameter that significantly impacts the performance of FedGAI is the local GAN model update iterations n\_iter. As shown in Figure~\ref{fig:commun_n_iter}(a), FL algorithms converged approximately in the 40th communication round. Subsequently, it can be observed from Figure~\ref{fig:commun_n_iter}(b) that the FL algorithms usually converge in the 23rd communication round. Finally, the FL algorithm can be found through Figure~\ref{fig:commun_n_iter}(c) and (d) to converge around the 17th and the 12th communication rounds, respectively. It can be found that with the n\_iter increases, FL algorithms converge rapidly. It also can be observed in Figure~\ref{fig:commun_n_iter} (a-d) that FedGAI is basically the fastest of the four algorithms; in addition, the performance of the other three methods is relatively oscillatory, which is not as stable as FedGAI.

\subsubsection{FL Performance on Scalable Clients.}
\begin{figure}[t]
    \centering
    \begin{subfigure}[b]{0.49\textwidth}
        \centering
        \includegraphics[width=\columnwidth]{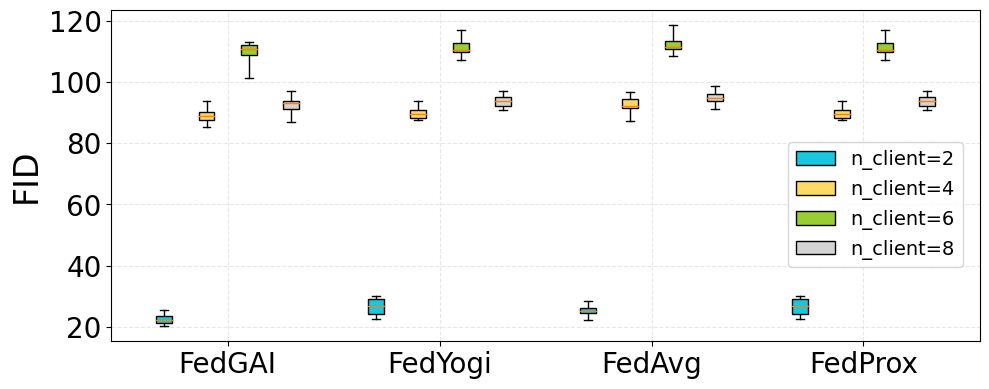}
        \caption{\small Average FID}
        \label{fig:sub1}
    \end{subfigure}
    \begin{subfigure}[b]{0.49\textwidth}
        \centering
        \includegraphics[width=\columnwidth]{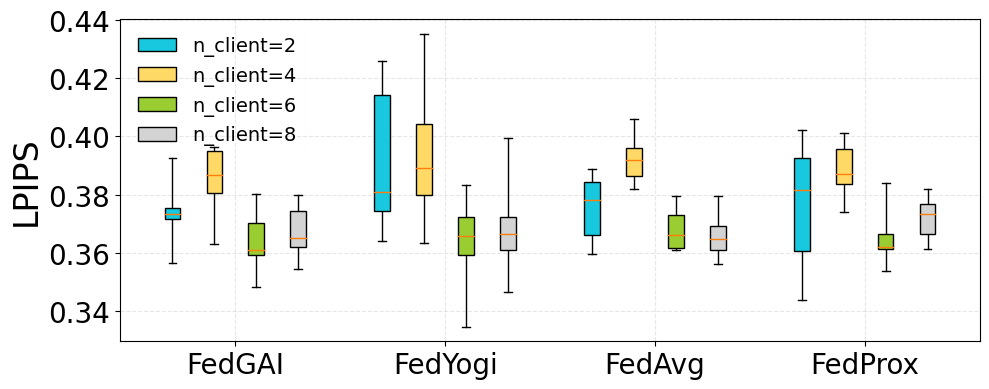}
        \caption{\small Average LPIPS}
        \label{fig:sub2}
    \end{subfigure}
    \begin{subfigure}[b]{0.49\textwidth}
        \centering
        \includegraphics[width=\columnwidth]{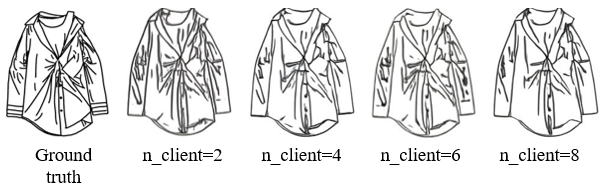}
        \caption{\small Corresonding sketches of FedGAI}
        \label{fig:sub3}
    \end{subfigure}
    \caption{The impact of client number on FL.}
    \label{fig:robustness of n_client}
\end{figure}

The number of clients engaged in federated learning directly influences the convergence of model aggregation, relevant to system robustness and scalability. On one hand, if only a limited number of clients participate in federated learning, the model may diverge due to insufficient data. On the other hand, involving a large number of clients in federated learning could lead to traffic congestion or introduce divergent client models. Selecting appropriate number of clients is an intriguing and challenging task. Figure~\ref{fig:robustness of n_client} presents our FL robustness and scalability evaluations across varying numbers of designers. As shown in  Figure~\ref{fig:robustness of n_client} (a), it is evident that the optimal performance is achieved when only two clients utilize FL for style fusion, as indicated by the smallest FID value. Furthermore, FedGAI consistently exhibits the lowest FID across different numbers of client conditions, demonstrating its scalability. Turning to Figure~\ref{fig:robustness of n_client} (b), it is apparent that the LPIPS interval range of FedGAI is narrower compared to other algorithms, indirectly affirming the robustness of FedGAI under different client scenarios. Notably, FedYogi's oscillatory performance has temporarily excluded it from the comparison range. In this context, FedGAI consistently exhibits the lowest LPIPS across various numbers of client conditions. Figure~\ref{fig:robustness of n_client} (c) includes hand-drawn sketches as well as generated sketches with FedGAI style fusion from different number of designers (n\_clients) is 2, 4, 6, and 8, respectively.

\section{Related Work}\label{related work}
\textbf{Generative AI Tools for Design.} In recent years, Generative AI enabled the automated generation of substantial content within limited timeframes, as opposed to content authored by humans~\cite{cao2023comprehensive}. In the realm of image generation, recent advancements have seen the emergence of autoregressive Transformer-based models like DALL-E~\cite{pmlr-v139-ramesh21a} and Parti~\cite{yu2022scaling}, as well as diffusion-based models like Imagen~\cite{saharia2022photorealistic} and Stable Diffusion~\cite{rombach2022high}. The diffusion model empowers GAI in several key aspects: Personalized Image Generation: This pertains to the capacity of GAI to produce images tailored to individual preferences or requirements. It encompasses the creation of custom avatars, profile pictures, or artwork aligning with a user's distinct style~\cite{ruiz2023dreambooth,hu2021lora}.
Image Controllable Generation and Editing: AI models offer fine-grained control over generated images, permitting users to manipulate color, style, content, and composition. This level of control facilitates precise image creation, particularly beneficial in graphic design and digital art~\cite{brooks2023instructpix2pix,yang2023paint}.
Video Controllable Generation and Editing: Extending image control to video content, Generative AI can manipulate video sequences, apply artistic filters, edit scenes, or even craft entirely new video content based on user input. This technology finds utility in video production, special effects, and personalized content creation~\cite{singer2022make,zhao2023controlvideo}. Three-Dimensional Scene Generation: This entails crafting 3D environments, objects, or scenes using GAI. AI models can generate realistic 3D scenes, objects, or landscapes, simplifying the creation of immersive digital experiences for designers and developers.

\vspace{1.5mm}

\noindent{\textbf{Collaborative Design and Style Fusion.}} Collaboration has the potential to ignite creativity and innovation among diverse designers\cite{zamenopoulos2018co}. Joint creations and brand collaborations~\cite{mrad2019karl} serve as collaborative efforts that contribute to mutual inspiration, uniting a range of ideas, styles, and visual elements. Designers draw inspiration from each other's design styles, resulting in the amalgamation of various aesthetics, cultures, and experiences. 
However, collaborative design is not without its challenges, including communication and understanding barriers\cite{maier2021factors}, issues related to creative protection, and constraints on time and resources\cite{wang2023mean}. In the realm of collaborative design, sketches assume a crucial role by expediting the initial creation of design prototypes. They also function as a communication medium, facilitating the sharing and integration of creative ideas among designers. With the emergence and advancement of Generative Adversarial Networks (GANs)~\cite{goodfellow2014generative}, deep learning-based sketch generation algorithms have gained prominence. For instance, Ge et al. introduced DoodlerGAN, tailored for creative sketch generation within the Birds and Creatures category~\cite{ge2020creative}. Liu et al. presented the Tom model, specialized in generating landscape-related sketches~\cite{liu2021self}. More recently, Wang et al. introduced SketchKnitter, utilizing the Diffusion model to achieve vectorized sketch generation by learning the data distribution of stroke point positions and stroke states from real human sketches~\cite{wang2022sketchknitter}. However, these approaches primarily focus on converting images into sketches, overlooking a crucial aspect of design, that is individual designer's distinct design styles. In response to this gap, Wu et al. pioneered the concept of generating designer-style models~\cite{wu2023styleme}. However, their approach entails a significant number of model parameters, which limits its versatility and practicality across various scenarios.

\section{Discussion}\label{discussion}
\noindent{\textbf{Impact on Generative AI.}} 
Rarely have studies addressed scenarios involving Generative AI in a collaborative design environment, Our FedGAI generative AI framework provides specific task allocation and coordination between cloud and edge to support the scalable capability of different users. FedGAI employs federated learning, which could effectively protect the personal data of different designers, thus eliminating the privacy leakage of designers in the collaboration process. A light GAN for sketch generation was proposed to balance the generation quality and computing resource constraints on edge devices. Furthermore, we integrated GAN compression for local model acceleration, meanwhile alleviating the communication overhead of cloud-edge collaboration within federated learning. Through FedGAI, designers can produce sketches that amalgamate diverse design styles contributed by their peers, thereby drawing inspiration from collaborative efforts without necessitating data disclosure or uploading.

\noindent{\textbf{Generality of FedGAI.}}
We first evaluate the sketch generation capabilities of FedGAI by comparative and ablation study, which prove that FedGAI can produce multi-styled sketches of comparable quality to human-designed ones and significantly enhance efficiency in creative design. Moreover, to investigate the superiority of our FedGAI, we integrated three commonly used FL algorithms with the GAN model in FedGAI and employed them as baseline comparisons in terms of sketch generation quality, sketch generation stability, and sketch generation speed. The result shows that FedGAI outperforms other baselines in three metrics, which mostly benefit from our personalized client selection strategy and FL training strategy. These approaches also can be transferred into other similar fields, providing a new reliable paradigm of generative AI-based collaboration.

\noindent{\textbf{Limitations and Future Work.}}
Although current FedGAI has achieved style fusion among designers while protecting the privacy of their data, its user threshold is still not relatively low for client users who lack deep learning knowledge. While the efficiency and effectiveness of the algorithm have been verified, there exists still a gap between the quality of sketches generated by the algorithm and those drawn by hand. Future work will focus on further research and improvement of the generation model, increasing the diversity of training data to produce higher-quality and more diverse fashion design works. We will also explore the integration of additional design elements into the system, to make the system more comprehensive.

\section{CONCLUSION}\label{conclusion}
FedGAI is a federated generative AI fashion design system capable of autonomously producing sketches with designer-style characteristics using limited computational resources and storage space. Simultaneously, it accomplishes style fusion among designers while ensuring the confidentiality of their data. To the best of our knowledge, it stands as the first fashion design creative support tool to support copyright protection. It comprises components based on GAN models, GAN compression techniques, and federated learning algorithms, among others. We also conducted experiments to evaluate the system, and the results show that the system performs well in terms of efficiency, effectiveness, scalability, and robustness compared to other benchmarks.

\bibliographystyle{IEEEtran}

\bibliography{ton.bib}

\end{document}